\documentclass[lettersize,journal]{IEEEtran}
\usepackage{amsmath,amsfonts}
\usepackage{algorithmic}
\usepackage{algorithm}
\usepackage{array}
\usepackage[caption=false,font=normalsize,labelfont=sf,textfont=sf]{subfig}
\usepackage{textcomp}
\usepackage{stfloats}
\usepackage{url}
\usepackage{verbatim}
\usepackage{graphicx}
\usepackage[table]{xcolor}
\definecolor{myblue}{RGB}{46,48,146}
\usepackage{hyperref}
\hypersetup{
    colorlinks=true,
    linkcolor=black!50!red,
    urlcolor=black!50!red,
    citecolor=black!50!blue
}
\usepackage{booktabs}
\usepackage{multirow}
\usepackage{graphicx}

\definecolor{lightgraycell}{HTML}{E6E6E6}

\begin{document}

\title{HyperbolicRAG: Enhancing Retrieval-Augmented Generation with Hyperbolic Representations}

\author{Linxiao Cao,~\IEEEmembership{Student Member,~IEEE,} Ruitao Wang,~\IEEEmembership{Student Member,~IEEE,} Jindong Li,~\IEEEmembership{Student Member,~IEEE,} Zhipeng Zhou,~\IEEEmembership{Member,~IEEE,} Menglin Yang

        % <-this % stops a space
\thanks{Linxiao Cao, Ruitao Wang, Jindong Li and Menglin Yang are with Hong Kong University of Science and Technology (Guangzhou), Guangzhou, China. Email: \{lcao950, rwang356, jli839\}@connect.hkust-gz.edu.cn,
menglin.yang@outlook.com. }
\thanks{Zhipeng Zhou is with Nanyang Technological University, Singapore. Email: zzpustcml@gmail.com}
\thanks{\textit{Corresponding author: Menglin Yang}}
% <-this % stops a space
}

\markboth{}%
{Shell \MakeLowercase{\textit{et al.}}: A Sample Article Using IEEEtran.cls for IEEE Journals}


\maketitle

\begin{abstract}
Retrieval-augmented generation (RAG) enables large language models (LLMs) to access external knowledge, helping mitigate hallucinations and enhance domain-specific expertise. 
% Current methods primarily capture surface-level semantic similarity rather than provide sufficient support for reasoning that spans multiple documents.
Graph-based RAG enhances structural reasoning by introducing explicit relational organization that enables information propagation across semantically connected text units. However, these methods typically rely on Euclidean embeddings that capture semantic similarity but lack a geometric notion of hierarchical depth, limiting their ability to represent abstraction relationships inherent in complex knowledge graphs.
To capture both fine-grained semantics and global hierarchy, we propose HyperbolicRAG, a retrieval framework that integrates hyperbolic geometry into graph-based RAG. 
% The framework learns depth-aware representations within a shared Poincaré manifold, aligning semantic similarity with hierarchical containment. Furthermore, an unsupervised contrastive regularization further enforces geometric consistency across abstraction levels.
% At inference time, retrieval signals from Euclidean and hyperbolic spaces are jointly exploited through a mutual-ranking fusion mechanism that emphasizes cross-space agreement. 
HyperbolicRAG introduces three key designs: (1) a depth-aware representation learner that embeds nodes within a shared Poincaré manifold to align semantic similarity with hierarchical containment, (2) an unsupervised contrastive regularization that enforces geometric consistency across abstraction levels, and (3) a mutual-ranking fusion mechanism that jointly exploits retrieval signals from Euclidean and hyperbolic spaces, emphasizing cross-space agreement during inference.
Extensive experiments across multiple QA benchmarks demonstrate that HyperbolicRAG outperforms competitive baselines, including both standard RAG and graph-augmented baselines.
\end{abstract}


\section{Introduction}
\IEEEPARstart{L}{arge} language models (LLMs) have demonstrated remarkable capabilities across a wide range of natural language processing tasks, including question answering, summarization, dialogue generation, and personalization~\cite{yi2024survey,zhang2024comprehensive,liu2025survey}.
Despite their strong generalization ability, LLMs inevitably suffer from knowledge staleness and hallucination, as their internal parameters cannot be easily updated with newly emerging facts or domain-specific information~\cite{fan2024survey}.

To mitigate these limitations, retrieval-augmented generation (RAG)~\cite{lewis2020retrieval} has emerged as a powerful paradigm that equips LLMs with access to external knowledge bases. By retrieving relevant documents at inference time and conditioning generation on this evidence, RAG systems can provide more up-to-date and contextually grounded responses, thereby reducing reliance on outdated or incomplete parametric knowledge.
% In a typical RAG pipeline, a retriever first selects relevant documents from a large corpus, and a generator then produces responses conditioned on both the query and the retrieved evidence. This approach has significantly improved factual accuracy and adaptability across open-domain tasks.
% Nevertheless, existing retrieval mechanisms remain limited in their reasoning capacity. 
% Conventional dense retrievers\footnote{Dense retrievers encode queries and documents into continuous vector representations and retrieve results based on embedding similarity~\cite{karpukhin2020dense}.} emphasize local semantic matching rather than relational or hierarchical understanding, making it difficult to perform multi-hop reasoning or to capture the structural organization of knowledge.
% To mitigate these limitations, recent studies, 
Building on this idea, graph-based RAG methods, 
such as G-Retriever~\cite{he2024g}, GraphRAG~\cite{edge2024local}, LightRAG~\cite{guo2024lightrag}, HippoRAG~\cite{jimenez2024hipporag} and HippoRAG2~\cite{gutierrez2025rag}, have organized the retrieved or corpus-level documents into graph structures. 
In these approaches, documents, entities, and concepts are represented as interconnected nodes linked by semantic or relational edges, enabling more structured access to knowledge. 
This paradigm enables multi-hop evidence aggregation through explicit graph traversal or message passing, thereby improving reasoning over linked knowledge. 

However, graph-based retrieval and reasoning methods typically embed nodes in flat Euclidean spaces, which are ill-suited for representing the hierarchical dependencies that underlie complex knowledge~\cite{krioukov2010hyperbolic,pan2021hyperbolic}.
Consider the query, \textit{``How does long-term tension (chronic stress) lead to weakened immunity?"} A standard dense retriever\footnote{Dense retrievers encode queries and documents into continuous vector representations and retrieve results based on embedding similarity~\cite{karpukhin2020dense}.} usually returns passages about broad themes like ``health" or ``stress," which are superficially relevant yet too generic to reflect the underlying mechanisms.
This behavior arises from the hubness inherent in high-dimensional Euclidean embedding spaces~\cite{tomasev2013role,radovanovic2010hubs}: semantically broad concepts occupy central regions that lie close to many queries, causing retrieval to disproportionately favor high-frequency, generic nodes.
Consequently, graph traversal treats nodes as if they lie on a single semantic plane, overlooking that ``cortisol release” is a specific descendant of ``stress” along a causal and ontological hierarchy. 
% as lying on the same semantic plane, failing to recognize that “cortisol release” is a more specific descendant of “stress”. 
In other words, graph-based propagation can connect entities while remaining largely insensitive to the hierarchical geometry that structures complex domains.

\begin{figure*}[htbp]
    \centering
    \includegraphics[width=\linewidth]{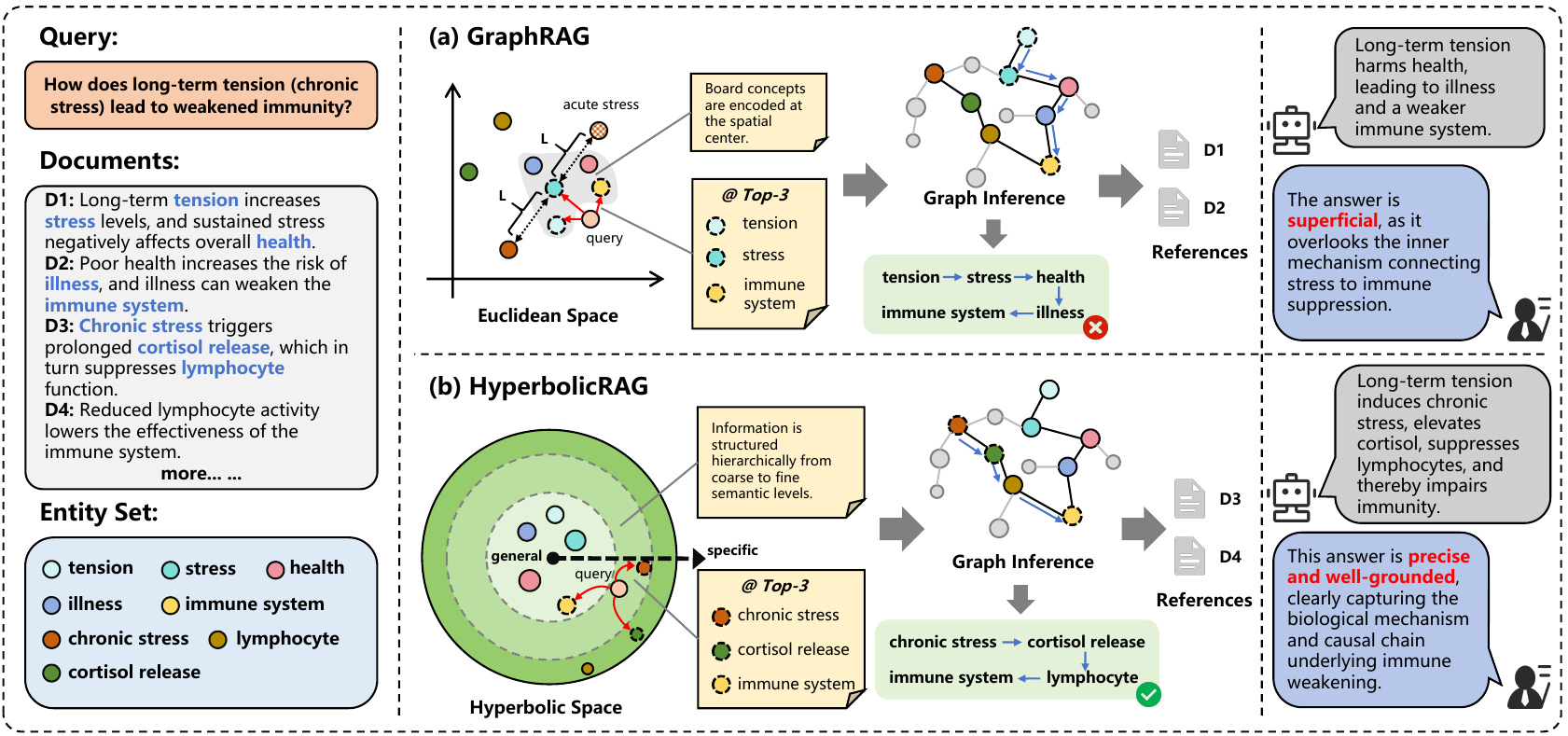}
    \caption{
    % \textbf{Comparison of Euclidean vs. hyperbolic embedding effects on retrieval-augmented multi-hop reasoning.} (a) In Euclidean space, embeddings primarily reflect surface-level semantic similarity. High-level concepts (e.g., stress) act as semantic hubs because their vectors approximate the mean of multiple subordinate contexts (e.g., acute stress, chronic stress). Linear distance metrics therefore make these hub nodes geometrically close to many queries, causing top-k activation to favor broad concepts and leading graph propagation (eg., PPR) to spread over general subgraphs. The resulting answers tend to be superficial. (b) In hyperbolic space, hierarchical depth is encoded radially: general concepts concentrate near the center while fine-grained mechanistic facts align near the boundary. A mechanism-oriented query is mapped closer to boundary nodes (via depth alignment), which yields initial activation on specific mechanism nodes (e.g., chronic stress, cortisol release, lymphocyte). Subsequent propagation remains focused within the mechanism subbranch, producing more precise, causal answers.
    Comparison of Euclidean and hyperbolic embedding effects on retrieval-augmented multi-hop reasoning.
(a) In Euclidean space, embeddings reflect surface-level similarity. General concepts (e.g., stress) act as semantic hubs, making top-$k$ retrieval and graph propagation drift toward broad, generic subgraphs.
(b) In hyperbolic space, hierarchical depth is radially encoded: abstract nodes lie near the center, while specific facts align near the boundary. Queries are thus aligned to relevant mechanism nodes (e.g., chronic stress, cortisol release), yielding more precise and causally focused reasoning.}
    \label{fig:intro}
\end{figure*}

To address these challenges, and inspired by findings that human perception structures concepts in tree-like hierarchies where general concepts subsume more specific sub-concepts~\cite{zhang2023hippocampal}, we propose integrating hyperbolic geometry into GraphRAG.
% Such structures are naturally modeled in hyperbolic geometry, where radial distance encodes semantic depth and containment with minimal distortion~\cite{sarkar2011low,ganea2018hyperbolic_entail}. 
Hyperbolic geometry naturally models such structures by encoding semantic depth and containment with minimal distortion~\cite{sarkar2011low,ganea2018hyperbolic_entail}: radial distance represents levels of specificity, and the exponential expansion of hyperbolic space accommodates large and deep hierarchies.
As illustrated in Fig.~\ref{fig:intro}, in Euclidean space (top), general and specific concepts co-locate on a flat surface, limiting the separation of leaf-leaf nodes and blurring hierarchical boundaries. 
In contrast, within hyperbolic space (bottom), general concepts are positioned near the center, while specific facts are located toward the boundary. This arrangement exploits the exponential growth property to preserve hierarchical containment relations.

% Recent advances such as GraphRAG~\cite{edge2024local} and HippoRAG~\cite{jimenez2024hipporag} attempt to introduce structural awareness by constructing heterogeneous graphs.
% While these models improve connectivity and enable relevance propagation, their embeddings remain Euclidean and flat.
% As a result, graph traversal still treats all nodes as lying on the same semantic plane—without understanding that “cortisol release” is a more specific descendant of “stress”. In other words, they propagate across connections but remain blind to the hierarchical geometry of knowledge. This gap motivates our work. Human knowledge is inherently tree-like, where general concepts organize increasingly specific sub-concepts. Such structures are naturally modeled in hyperbolic geometry, where radial distance encodes depth and containment with minimal distortion. Figure \ref{fig:intro} illustrates this geometric limitation: in the Euclidean space (left), general and specific concepts co-locate on a flat surface, blurring hierarchical boundaries. In contrast, the hyperbolic space (right) embeds general concepts near the center and specific facts toward the boundary, preserving containment relations through radial distance. 

Building upon this geometric insight, we develop HyperbolicRAG, a hierarchy-aware retrieval framework that integrates hyperbolic geometry into graph-based RAG. 
% HyperbolicRAG encodes textual units and relational structures within a shared Poincaré manifold, allowing representations to simultaneously capture semantic proximity and hierarchical depth.
HyperbolicRAG introduces three key components. First, it predicts a semantic depth for each textual unit and projects it into a shared Poincaré manifold, where radial distance explicitly encodes hierarchical specificity. Second, a bidirectional alignment loss enforces consistent containment relations between passages and fine-grained factual evidence. Third, at inference time, HyperbolicRAG performs retrieval jointly in Euclidean and hyperbolic spaces and fuses their rankings, thereby balancing local semantic similarity with hierarchical relevance.

Our main contributions are summarized as follows:
\begin{itemize}
\item \textbf{Hierarchy-aware hyperbolic Representation.} We propose a hierarchy-aware retrieval framework that predicts a scalar depth for each textual unit and performs depth-controlled projection into a shared Poincaré manifold, preserving local semantics while encoding hierarchical structure via radial positions.
\item \textbf{Bidirectional containment alignment.} We introduce a bidirectional margin-based loss that aligns passages and facts, enabling the model to internalize containment relations across different granularities.
\item \textbf{Dual-space retrieval fusion.} We develop a dual-space retrieval mechanism that fuses Euclidean and hyperbolic reasoning via mutual-ranking fusion, improving robustness against noisy or overly generic evidence.
\item \textbf{Competitive results.} Extensive experiments on multiple QA benchmarks demonstrate that HyperbolicRAG consistently outperforms standard RAG and graph-augmented baselines, particularly on multi-hop reasoning tasks.
\end{itemize}

The remainder of this paper is organized as follows. 
Section~\ref{related work} reviews related work on retrieval-augmented generation, graph-based retrieval, and hyperbolic representation learning. 
Section~\ref{preliminary} summarizes the necessary background on hyperbolic geometry. 
Section~\ref{method} details the proposed framework. 
Section~\ref{experiments} presents experimental results and analysis. 
Finally, Section~\ref{conclusion} concludes the paper.

\section{Related Work}\label{related work}
\subsection{Retrieval Augment Generation}
% RAG has become a central paradigm for overcoming the static nature of LLMs by decoupling parametric knowledge from dynamic information access. In its standard form, RAG relies on dense vector retrieval to fetch relevant passages, which are then injected into the model’s context window for generation. This approach has proven effective in factual recall tasks, where local semantic similarity suffices to identify useful evidence. However, when applied to large-scale, unstructured corpora—such as scientific papers, textbooks, or technical reports—RAG faces substantial challenges. Domain knowledge is often scattered across documents with varying accuracy and completeness, lacking explicit hierarchical organization between concepts. As a result, conventional chunk-based indexing, though efficient for retrieval, tends to sacrifice long-range contextual coherence, limiting the system’s ability to support complex reasoning and multi-hop inference.

RAG has emerged as a key paradigm to overcome the static knowledge limitation of LLMs, decoupling parametric knowledge stored in model weights from dynamic retrieval over external corpora \cite{lewis2020retrieval}. In its original form, RAG relies on dense vector retrieval to fetch topically relevant passages, which are then injected into the LLM context to ground generation. 
% This approach performs well on single-hop factual recall tasks, where local semantic similarity suffices.
% However, vanilla RAG faces challenges in large-scale, unstructured corpora such as scientific literature or multi-chapter textbooks, where domain knowledge is often fragmented across passages with varying granularity and completeness. This fragmentation limits its ability to support multi-hop reasoning and hierarchical query answering.
% To address these limitations, 
Recent research has explored structured extensions of RAG, with graph-based retrieval emerging as a prominent direction \cite{zhang2025survey,peng2024graph,procko2024graph}. GraphRAG methods organize textual units into graphs, where edges encode semantic relationships such as entity mentions or inter-entity associations. This structured representation enables relevance propagation across connected nodes, thereby enriching contextual grounding for downstream LLM reasoning.

Despite sharing a common principle, existing GraphRAG variants diverge substantially in their design focus.
Hierarchical or community-based approaches, such as Microsoft GraphRAG~\cite{edge2024local} and LazyGraphRAG~\cite{lazy2024graph}, construct multi-level partitions of the corpus (e.g., chapter → section → subsection) to support both local retrieval within communities and global retrieval across them, thereby balancing retrieval precision and corpus coverage.
Another line of work centers on structure-optimized designs. LightRAG~\cite{guo2024lightrag} enriches retrieval by introducing graph-enhanced indexing that combines entity–relation graph construction with key–value profiling, together with a dual-level retrieval strategy to integrate entity-level accuracy with topic-level breadth. GRAG~\cite{hu2024grag} similarly focuses on structural optimization but emphasizes robustness: it employs soft pruning to suppress irrelevant nodes and incorporates graph-aware prompt tuning to reduce retrieval noise.
A third direction explores task-adaptive frameworks. StructRAG~\cite{li2024structrag} dynamically incorporates multiple relation types (e.g., part-of, causal) to better support complex reasoning demands, whereas KAG~\cite{liang2025kag} relies on human-curated schemas to construct high-precision, domain-specific knowledge graphs that surpass fully automated extraction pipelines in specialized settings.
Yet, these structured RAG variants ignore the underlying geometry where the node representation relies on Euclidean embeddings only, implying that even though nodes are connected relationally, their spatial representation fails to encode hierarchical containment. 
This geometric limitation motivates us to incorporate hyperbolic geometry into the retrieval process.
% In HyperbolicRAG, knowledge representations are learned within a curved space where the negative curvature provides a natural inductive bias for encoding hierarchical relations.

% To address these limitations, recent research has explored structured extensions of RAG, with a particular emphasis on graph-based approaches \cite{zhang2025survey,peng2024graph,procko2024graph}. GraphRAG methods aim to organize textual information into structured graphs that capture entities, relations, or hierarchical communities, thereby providing richer retrieval contexts. Early efforts such as Microsoft GraphRAG \cite{edge2024local}  and LazyGraphRAG \cite{lazy2024graph} employ hierarchical community search to combine local and global querying for more comprehensive responses. LightRAG \cite{guo2024lightrag} introduces dual-level retrieval and graph-enhanced indexing to improve scalability, while GRAG \cite{jin2024ragcache} incorporates soft pruning of irrelevant nodes and graph-aware prompt tuning to mitigate noise in subgraph retrieval. StructRAG \cite{li2024structrag} further adapts graph schemas dynamically to task-specific needs, and KAG \cite{liang2025kag} leverages semantic reasoning and human-curated schemas to construct domain-specific knowledge graphs with reduced noise compared to automatic OpenIE pipelines. These strategies significantly enhance retrieval precision and contextual depth, enabling LLMs to perform better on discourse-level comprehension and multi-hop queries.

\subsection{Hyperbolic Representation Learning}
Hyperbolic geometry has become an effective paradigm for modeling hierarchical data, addressing a key limitation of Euclidean spaces: their difficulty in embedding tree-like or nested structures with low distortion~\cite{sarkar2011low}. 
Early foundational works such as Poincaré and Lorentz embeddings~\cite{nickel2017poincare,nickel2018learning} and hyperbolic neural networks~\cite{ganea2018hyperbolic,chami2019hyperbolic,dai2021hyperbolic,shimizu2020hyperbolic,chen2022fully} demonstrate that hyperbolic manifolds provide exponentially expanding representational capacity that naturally aligns with hierarchical and scale-free structures. 
Besides, hyperbolic geometry has also been applied with hyperbolic metric learning~\cite{yan2021unsupervised,ermolov2022hyperbolic,dhingra2018embedding,lang2022hyperbolic}, and graph learning, like HyperIMBA~\cite{fu2023hyperbolic}, HVGNN~\cite{sun2021hyperbolic},  H2H-GCN~\cite{dai2021hyperbolic}, H2SeqRec~\cite{li2021hyperbolic}. 
% Its utility has further been demonstrated in hyperbolic graph learning, where HyperIMBA~\cite{fu2023hyperbolic} addresses hierarchy imbalance in node classification, HVGNN~\cite{sun2021hyperbolic} models temporal evolution and uncertainty in dynamic graphs, H2H-GCN~\cite{dai2021hyperbolic} performs message passing directly on curved manifolds to preserve global structure, and H2SeqRec~\cite{li2021hyperbolic} captures higher-order dependencies through hyperbolic hypergraph convolutions. 
These empirical advances align with theoretical results showing that hyperbolic spaces enable low-distortion tree embeddings~\cite{sarkar2011low}, exhibit favorable generalization behavior for hierarchical data~\cite{suzuki2021generalization}, and provide high representational efficiency for complex structures~\cite{mishne2023numerical}.

Recent advances are pushing the frontier of hyperbolic geometry into LLMs. Yang et al.~\cite{yang2024hyperbolic} demonstrate that token embeddings exhibit inherent hyperbolicity and propose a hyperbolic adaptation for LLMs that enhances downstream reasoning performance. 
Desai et.al~\cite{desai2023hyperbolic} introduce hyperbolic geometry into image-text representation.
He et al.~\cite{he2025helm} propose curvature-adaptive hyperbolic LLM architectures and train hyperbolic LLMs from scratch. 
% Collectively, these studies highlight a growing convergence between hyperbolic modeling and large-scale language understanding.
Despite these advances, hyperbolic representation learning has rarely been applied to RAG. 
% While effective for hierarchical modeling, its potential to enhance multi-hop retrieval and structured evidence aggregation remains underexplored. 
Our work bridges this gap by integrating hyperbolic geometry into RAG through explicit regularization and dual-space retrieval.

\section{Preliminary}\label{preliminary}
% Hyperbolic geometry offers a natural space for representing hierarchical or tree-structured data with low distortion. Its negative curvature induces exponential volume growth, a property that aligns well with the branching patterns of real-world knowledge. 
There are several isometric hyperbolic models~\cite{he2025hyperbolic,peng2021hyperbolic,yang2022hyperbolicsurvey}, including the Poincaré ball model, the Lorentz model, Klein model, which show
different characteristics but are mathematically equivalent.
In this work, we employ the Poincaré ball model, which provides a conformal and analytically convenient formulation of hyperbolic space. 
Its closed-form geodesics and mappings integrate seamlessly with Euclidean neural encoders, and its radial coordinate furnishes an interpretable measure of hierarchical depth. These features make the Poincaré model well-suited to our depth-controlled projection mechanism and contribute to stable end-to-end training. 
Although we instantiate our method in the Poincaré ball, it is also compatible with other hyperbolic models. The core geometric concepts of Poincaré ball used in our framework are summarized below.

\textbf{Poincaré Ball Model.} The $d$-dimensional Poincaré ball with negative curvature $- c$ ($c>0$) is defined as
$
    \mathbb{H}^c_d = \{\, \mathbf{x} \in \mathbb{R}_d : c\|\mathbf{x}\|^{2} < 1 \,\}.
$
It is a conformal model in which angles are preserved and all points lie within a Euclidean ball of radius $1/\sqrt{c}$. The Riemannian metric is $
    g_{\mathbf{x}} = \lambda_{\mathbf{x}}^{2} g^{E},\text{ and }
    \lambda_{\mathbf{x}}^c = \frac{2}{1 - c\|\mathbf{x}\|^{2}},
$
where $g^{E}$ denotes the Euclidean metric. The conformal factor grows rapidly near the boundary, producing the characteristic expansion of hyperbolic space and enabling compact representation of deeply nested hierarchical structures.

\textbf{Geodesic and Radial Distances.}
Distances in the Poincaré ball follow the induced Riemannian geometry. For points
$\mathbf{u}, \mathbf{v} \in \mathbb{H}^c_d$, the hyperbolic geodesic distance is
\begin{equation}
    d_{\mathbb{H}}^c(\mathbf{u}, \mathbf{v})
    = \frac{1}{\sqrt{c}}
      \operatorname{arcosh}\!\left(
        1 + 2c\,
        \frac{\|\mathbf{u}-\mathbf{v}\|^{2}}
             {(1 - c\|\mathbf{u}\|^{2})(1 - c\|\mathbf{v}\|^{2})}
      \right).
\end{equation}
This distance reflects both the Euclidean separation of the points and their proximity to the boundary, which causes the metric to expand and naturally induces hierarchical organization: points near the origin correspond to more general concepts, whereas points near the boundary denote more specific ones.
The radial distance from the origin is
\begin{equation}
    d_{\mathbb{H}}^c(\mathbf{x}, \mathbf{0})
    = \frac{1}{\sqrt{c}}
      \operatorname{arcosh}\!\left(
        1 + 2c\,
        \frac{\|\mathbf{x}\|^{2}}
             {1 - c\|\mathbf{x}\|^{2}}
      \right),
\end{equation}
which provides a direct geometric measure of hierarchical depth in the embedding space.

\textbf{Exponential and Logarithmic Maps.}
To couple hyperbolic representations with Euclidean neural encoders, we use the mappings between the tangent space at the origin and the manifold. The tangent space provides a locally Euclidean parameterization that allows standard neural operations to interface with hyperbolic geometry.

The exponential map sends a tangent vector $\mathbf{v} \in T_{\mathbf{0}}\mathbb{H}^c_d$ onto the manifold:
\begin{equation}
    \exp_{\mathbf{0}}^c(\mathbf{v})
    = \tanh\!\left(\sqrt{c}\,\|\mathbf{v}\|\right)
      \frac{\mathbf{v}}{\sqrt{c}\,\|\mathbf{v}\|}.
\end{equation}

Its inverse, the logarithmic map, returns a point $\mathbf{u} \in \mathbb{H}^c_d$ to the tangent space:
\begin{equation}
    \log_{\mathbf{0}}^c(\mathbf{u})
    = \frac{1}{\sqrt{c}}
      \tanh^{-1}\!\left(\sqrt{c}\,\|\mathbf{u}\|\right)
      \frac{\mathbf{u}}{\|\mathbf{u}\|}.
\end{equation}

Together, these maps provide a smooth interface between Euclidean and hyperbolic computations, enabling end-to-end training while preserving the hierarchical structure encoded by the manifold.

\begin{figure*}[htbp]
    \centering
    \includegraphics[width=\linewidth]{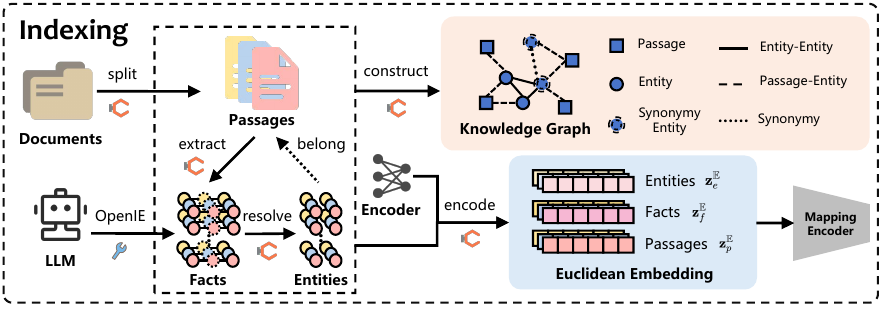}
    \caption{
    \textbf{Indexing pipeline.}
    Given a document collection, the framework first performs \emph{chunking} to obtain passages, from which an \emph{OpenIE extractor} derives relational triples and normalized entity mentions. 
    Passages, entities, and facts are then encoded into dense vectors using a pretrained encoder. 
    Finally, a heterogeneous knowledge graph is constructed by linking (i) entity--entity pairs co-occurring in triples, (ii) passage--entity pairs grounded in text, and (iii) synonymy links between semantically similar entities. 
    % The graph, together with cached embeddings, forms the structured index for subsequent hyperbolic projection and retrieval.
    }
    \label{fig:index}
\end{figure*}
\section{Methodology}\label{method}
\subsection{Overview}
\begin{figure*}
    \centering
    \includegraphics[width=\linewidth]{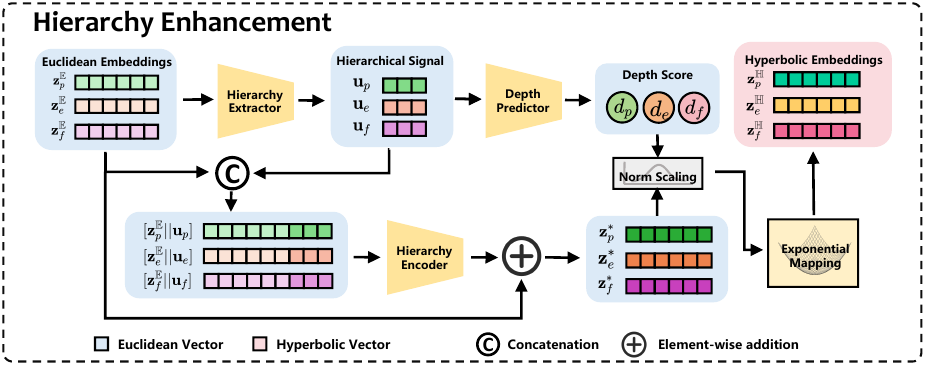}
    \caption{\textbf{Overview of the hierarchical enhancement process.} 
    Given Euclidean embeddings of passages, entities, and facts, the model first extracts a \textit{hierarchical signal} $\mathbf{u}_v$. This signal serves two roles: it is concatenated with the original semantic embedding $\mathbf{z}_v^{\mathbb{E}}$ to form an enhanced Euclidean representation enriched with hierarchical cues, and it is also used to predict a depth score $d_v$ that reflects the relative granularity of each node. The predicted depth then regulates a radial rescaling of the enhanced embedding, assigning smaller norms to more generic concepts and larger norms to more fine-grained evidence. Finally, the depth-aligned vectors are projected into the Poincaré ball via the exponential map, producing hyperbolic embeddings $\mathbf{z}_v^{\mathbb{H}}$ that jointly encode semantic similarity and hierarchical structure.
    }
    \label{fig:mapping}
\end{figure*}

Our framework presents a dual-space retrieval framework that seamlessly incorporates hierarchical structural knowledge into RAG. The pipeline is organized into three complementary stages: \textit{Indexing Process} (illustrated in Fig.~\ref{fig:index}), \textit{Hierarchical Enhancement} (illustrated in Fig.~\ref{fig:mapping}) and \textit{Dual-space Retrieval Process} (illustrated in Fig.~\ref{fig:retrieval}).

\subsection{Indexing Process}
The goal of the indexing phase is to transform the raw corpus $\mathcal{D} = \{D_1, D_2, \dots, D_N\}$ into a structured representation that supports semantic retrieval and hierarchical reasoning.
This is achieved by constructing a heterogeneous knowledge graph $\mathcal{G} = (\mathcal{V}, \mathcal{E}_{\text{edge}})$, where passages and entities constitute the node set and their relationships are encoded through extracted factual and contextual connections.
The overall process consists of four sequential stages:
(1) \emph{document chunking} to segment the corpus into coherent retrieval units,
(2) \emph{relational extraction} to identify canonical entities and their interrelations,
(3) \emph{representation learning} to obtain semantic embeddings as node features, and
(4) \emph{graph construction} to define the topological structure of $\mathcal{G}$.

\subsubsection{Document chunking}
Each document $D_i$ is divided into shorter, semantically coherent segments, referred to as \emph{passages}.
This segmentation balances retrieval granularity and computational efficiency: passages should preserve sufficient local context for meaningful retrieval while avoiding unnecessary redundancy.
% Formally,
% $
% D_i \longrightarrow \{p_{i,1}, p_{i,2}, \dots, p_{i,M_i}\}, \quad p_{i,j} \in \mathcal{P},
% $
% where $p_{i,j}$ denotes the $j$-th passage from document $D_i$, $M_i$ the number of passages derived from $D_i$, and $\mathcal{P}$ the global set of all passages.
Formally, each document $D_i$ is decomposed into passages 
$D_i \to \{p_{i,1}, p_{i,2}, \dots, p_{i,M_i}\}$ with $p_{i,j} \in \mathcal{P}$, 
where $p_{i,j}$ denotes the $j$-th passage extracted from $D_i$, $M_i$ is the number of passages derived from $D_i$, and $\mathcal{P}$ denotes the global set of all passages.
Each passage later becomes a textual node in $\mathcal{G}$, serving as the retrieval unit to which entities and relational evidence are anchored.

\subsubsection{Relational extraction}  
To build a relationally grounded view of the corpus, we employ a two-stage extraction pipeline guided by an LLM.  
First, the model identifies salient entities from each passage $p \in \mathcal{P}$ to establish the entity context.  
Conditioned on these entities, it then extracts relational triples that describe their interactions:
\begin{equation}
\mathcal{T}(p) = \{ (e_s, r, e_o) \mid e_s, e_o \in \text{Entities}(p), \ r \in \text{Relations}(p) \}.
\end{equation}
Each extracted subject or object entity $(e_s, e_o)$ is incorporated into the global entity set $\mathcal{E}$ if not already present,  
and every triple $(e_s, r, e_o)$ is recorded as a fact $f \in \mathcal{F}$.
Facts are not represented as graph nodes; instead, they act as relational annotations linking passages and canonical entities.
Accordingly, the node set of the graph is $
\mathcal{V} = \mathcal{P} \cup \mathcal{E},
$
where passages and entities form the structural backbone of $\mathcal{G}$, while facts enrich their semantic connectivity.

\subsubsection{Representation learning}
After identifying the structural elements, we encode their semantic content into dense vector representations using a pretrained language model encoder, denoted $\mathrm{Enc}(\cdot)$.
These embeddings serve as \emph{node features} and provide the semantic foundation for later hierarchy-aware refinement.
Specifically, we obtain three types of embeddings:
\begin{align}
\mathbf{z}_p^\mathbb{E} &= \mathrm{Enc}(p), && p \in \mathcal{P},\ \nonumber \\
\mathbf{z}_e^\mathbb{E} &= \mathrm{Enc}(e), && e \in \mathcal{E},\ \nonumber \\
\mathbf{z}_f^\mathbb{E} &= \mathrm{Enc}(f), && f \in \mathcal{F}.
\end{align}
Passage embeddings $\mathbf{z}_p^\mathbb{E}$ capture contextual semantics at the text level;
entity embeddings $\mathbf{z}_e^\mathbb{E}$ encode concept-level meaning aggregated across mentions;
and fact embeddings $\mathbf{z}_f^\mathbb{E}$ represent relational semantics between entities.
Importantly, these embeddings do not define the graph topology; rather, they serve as semantic attributes that enable similarity computation and inform the subsequent hierarchy-aware embedding enhancement.

\subsubsection{Graph construction}  
The final step of indexing is to establish edges $\mathcal{E}_{\text{edge}}$ in $\mathcal{G}$ to encode complementary structural relationships between nodes. These edges are categorized into three types, each designed to capture a specific type of semantic connection:

\begin{itemize}
    \item \textbf{Entity–Entity edges}: Each fact triple $(s,r,o)$ (with $s$ and $o$ normalized to canonical entities $e_s, e_o \in \mathcal{E}$) induces an edge between $e_s$ and $e_o$. To quantify the strength of this relationship, we increment the edge weight by the co-occurrence frequency of $e_s$ and $e_o$ across all facts. Formally, the edge weight is updated as $w(e_s, e_o) \leftarrow w(e_s, e_o) + 1$, where $w(e_s, e_o)$ denotes the weight of the edge between $e_s$ and $e_o$. These edges capture local factual relations (e.g., “lung cancer” $\xrightarrow{\text{causes}}$ “chest pain”) and support the propagation of evidence among semantically related entities during retrieval.

    \item \textbf{Passage–Entity edges}: Each passage $p \in \mathcal{P}$ is connected to all entities $e \in \text{Entities}(p)$ (i.e., all entity mentions in $p$ after normalization). Formally, we add an edge $(p,e)$ to $\mathcal{E}_{\text{edge}}$ for every entity $e \in \mathrm{Entities}(p)$, i.e., $(p,e) \in \mathcal{E}_{\text{edge}}$ for all $e \in \mathrm{Entities}(p)$. These edges anchor entities to their original textual context and allow entity-level signals (such as relevance scores of query-related entities) to influence passage scoring, thereby enhancing the model’s ability to capture context-aware connections.

    \item \textbf{Synonymy edges}: To address lexical variability (i.e., different surface forms of the same entity, such as "U.S." vs. "United States" or "COVID-19" vs. "coronavirus disease 2019"), we connect entities whose embeddings exceed a predefined cosine similarity threshold $\tau_{\text{syn}}$. Formally, we add an edge $(e,e')$ to $\mathcal{E}_{\text{edge}}$ whenever $\cos(\mathbf{z}_e^{\mathbb{E}}, \mathbf{z}_{e'}^{\mathbb{E}}) \ge \tau_{\text{syn}}$, where $\cos(\cdot,\cdot)$ denotes cosine similarity and $e,e' \in \mathcal{E}$ are distinct entities. This construction strengthens connectivity among semantically equivalent entities expressed with different surface forms and alleviates graph fragmentation arising from lexical variation in heterogeneous graphs.

\end{itemize}

The resulting heterogeneous graph $\mathcal{G}=(\mathcal{V},\mathcal{E}_{\text{edge}})$, together with the cached embeddings $\{\mathbf{z}_p^\mathbb{E} \mid p \in \mathcal{P}\}, \{\mathbf{z}_e^\mathbb{E} \mid e \in \mathcal{E}\}, \{\mathbf{z}_f^\mathbb{E} \mid f \in \mathcal{F}\}$, forms a compact yet expressive index, supporting later hierarchy-aware projection and dual-space reasoning.

\subsection{Hierarchical Enhancement} \label{etoh}
The heterogeneous graph constructed during indexing captures explicit relational connectivity but remains geometrically flat. Euclidean embeddings encode local semantic similarity but fail to capture how broad passages encompass fine-grained facts. This flat geometry causes relevance propagation to drift toward generic or high-degree nodes, leading to unstable multi-hop reasoning and noisy retrieval.

We address this by refining node embeddings. Specifically, Euclidean embeddings of passages, entities, and facts are projected into a shared hyperbolic space $\mathbb{H}_d^c$, whose negative curvature naturally models tree-like hierarchies. Radial distance from the origin encodes hierarchical depth: general passages are placed near the center, while specific facts are pushed toward the boundary. 
A learned depth predictor assigns each element a scalar specificity score that determines its radial position. This hierarchy-aware embedding reshapes the query–node similarity distribution, focusing propagation on relevant, hierarchy-consistent subgraphs without modifying the underlying topology.

The following sections describe (1) the \textit{hyperbolic projection mechanism} and (2) \textit{unsupervised optimization} that enforce containment consistency between passages and facts.

\subsubsection{Hyperbolic projection}

Given Euclidean embeddings $\mathbf{z}_v^{\mathbb{E}}$ for passages, entities, and facts, 
we project them into the hyperbolic space $\mathbb{H}_d^c$ to obtain hierarchy-aware representations 
$\mathbf{z}_v^{\mathbb{H}}$. The projection process integrates semantic preservation with hierarchical control and proceeds as follows.

\paragraph{Hierarchy feature extraction}  
% While Euclidean embeddings $\mathbf{z}_v^{\mathbb{E}}$ encode topical similarity, 
% they lack cues that distinguish hierarchical granularity. 
% To capture such structure, we apply a nonlinear transformation 
% $\phi:\mathbb{E}^d \to \mathbb{E}^{d'}$:
% \begin{equation}
%     \mathbf{u}_v = \phi(\mathbf{z}_v^{\mathbb{E}}),
% \end{equation}
% where $\mathbf{u}_v$ extracts hierarchy-related features.
While Euclidean embeddings $\mathbf{z}_v^{\mathbb{E}}$ encode topical similarity, they lack cues that distinguish hierarchical granularity. To capture such structure, we firstly apply a nonlinear transformation $\phi:\mathbb{E}^d \to \mathbb{E}^{d'}$, yielding $\mathbf{u}_v=\phi(\mathbf{z}_v^{\mathbb{E}})$, where $\mathbf{u}_v$ captures hierarchy-related features.

\paragraph{Depth prediction}  
% A type-specific predictor ($\psi_{\text{pass}}$, $\psi_{\text{fact}}$, $\psi_{\text{ent}}$) 
% maps $\mathbf{u}_v$ to a scalar depth score $d_v \in [0,1]$:
% \begin{equation}
%     d_v = \psi_{\text{mode}(v)}(\mathbf{u}_v),
% \end{equation}
% where smaller $d_v$ indicates more general passages or entities, and larger $d_v$ 
% represents more specific facts. These depth scores later guide the radial positioning of nodes in the hyperbolic space.
% A type-specific predictor ($\psi_{\text{pass}}$, $\psi_{\text{fact}}$, $\psi_{\text{ent}}$) maps $\mathbf{u}_v$ to a scalar depth score $d_v\in[0,1]$, i.e., $d_v=\psi_{\text{mode}(v)}(\mathbf{u}_v)$, where smaller $d_v$ indicates more general passages or entities and larger $d_v$ corresponds to more specific facts. These depth scores later guide the radial placement of nodes in hyperbolic space.
A type-specific predictor ($\psi_{\text{pass}}$, $\psi_{\text{fact}}$, $\psi_{\text{ent}}$) maps the hierarchical signal $\mathbf{u}_v$ to a depth score $d_v\in[0,1]$ via $d_v=\psi_{\text{mode}(v)}(\mathbf{u}_v)$, where smaller scores correspond to more general nodes and larger scores to more specific ones. These depth values later determine the radial placement of nodes in the hyperbolic space.

\paragraph{Feature fusion with gating}  
To jointly preserve semantic and hierarchical information, we fuse 
$\mathbf{z}_v^{\mathbb{E}}$ and $\mathbf{u}_v$ via a gating mechanism:
\begin{align}
    \tilde{\mathbf{z}}_v^{\mathbb{E}} &= \rho([\mathbf{z}_v^{\mathbb{E}} \,\|\, \mathbf{u}_v]), \\
    \mathbf{m}_v &= \sigma(W_g \tilde{\mathbf{z}}_v^{\mathbb{E}}), \\
    \mathbf{z}_v^{*} &= \mathbf{m}_v \odot \mathbf{z}_v^{\mathbb{E}} + 
                       (1-\mathbf{m}_v)\odot \tilde{\mathbf{z}}_v^{\mathbb{E}},
\end{align}
where $\sigma(\cdot)$ is the element-wise sigmoid, $W_g\!\in\!\mathbb{E}^{d\times d}$ is the learnable weight matrix of the gating layer, and $\mathbf{m}_v\!\in\!(0,1)^d$ are per-dimension gates. The result $\mathbf{z}_v^{*}\!\in\!\mathbb{E}^{d}$ is the refined Euclidean embedding used in the subsequent depth-alignment step.

\paragraph{Radial depth alignment}
% To translate the predicted depth $d_v$ into spatial structure, 
% we regulate the $L_2$ norm of the refined embedding $\mathbf{z}_v^{*}$ as a linear function of $d_v$:
% \begin{equation}
%     \|\hat{\mathbf{z}}_v^{\mathbb{E}}\| = \alpha + \beta d_v,
% \end{equation}
% where $\alpha,\beta>0$ are hyperparameters satisfying $\alpha+\beta \le 1$, 
% ensuring all vectors remain within the Poincaré ball.
To translate the predicted depth $d_v$ into spatial structure, we regulate the $L_2$ norm of the refined embedding $\mathbf{z}_v^{*}$ by enforcing $\|\hat{\mathbf{z}}_v^{\mathbb{E}}\|=\alpha+\beta d_v$, where $\alpha,\beta>0$ and $\alpha+\beta\le 1$ to ensure all vectors remain inside the Poincaré ball.
The aligned embedding is then obtained by:
\begin{equation}
    \hat{\mathbf{z}}_v^{\mathbb{E}} = 
    \frac{\alpha + \beta d_v}{\|\mathbf{z}_v^{*}\|}\mathbf{z}_v^{*}.
\end{equation}
This step preserves semantic direction while encoding hierarchical depth as radial distance, where general nodes occupy inner regions and specific ones are placed closer to the boundary.

\paragraph{Mapping to hyperbolic space}
The aligned Euclidean vector $\hat{\mathbf{z}}_v^{\mathbb{E}}$ is then projected into the Poincaré ball 
$\mathbb{H}_d^{c} = \{\mathbf{x} \in \mathbb{E}^d \mid \|\mathbf{x}\| < 1\}$ 
via the exponential map at the origin:
% \begin{equation}
%     \mathbf{z}_v^{\mathbb{H}} = \exp_{\mathbf{0}}(\hat{\mathbf{z}}_v^{\mathbb{E}}) 
%     = \tanh(\|\hat{\mathbf{z}}_v^{\mathbb{E}}\|)\frac{\hat{\mathbf{z}}_v^{\mathbb{E}}}{\|\hat{\mathbf{z}}_v^{\mathbb{E}}\|}.
% \end{equation}
\begin{equation}
    \mathbf{z}_v^{\mathbb{H}} 
    = \exp_{\mathbf{0}}^{c}(\hat{\mathbf{z}}_v^{\mathbb{E}}). 
    % = \tanh\!\left(\frac{1}{2}\|\hat{\mathbf{z}}_v^{\mathbb{E}}\|\right)
    %   \frac{\hat{\mathbf{z}}_v^{\mathbb{E}}}{\|\hat{\mathbf{z}}_v^{\mathbb{E}}\|}.
\end{equation}
This mapping preserves the semantic direction of the Euclidean embedding while converting its norm into a radius on the Poincaré ball. 
Such curvature-aware embedding offers greater representational efficiency—capturing both local semantic similarity and global hierarchy structure.

\subsubsection{Unsupervised optimization}
The hyperbolic projection produces geometry-aware embeddings that encode hierarchical depth, yet it does not explicitly constrain the spatial relationships between passages and their contained facts. To enforce such containment consistency, we introduce a pair of margin-based contrastive objectives operating in both passage-to-fact and fact-to-passage directions.

\paragraph{Passage-to-Fact alignment}
For each passage $p \in \mathcal{P}$, let $\mathcal{F}(p)$ denote the set of facts extracted from it. Each positive pair $(p, f^+)$ is accompanied by a randomly sampled negative fact $f^-$ not associated with $p$. We encourage $p$ to be closer to $f^+$ than to $f^-$ in hyperbolic space by at least a margin $\gamma$:
\begin{align}
\mathcal{L}_{p \to f} = \sum_{f \in \mathcal{F}(p)}
\Big[ d_{\mathbb{H}}^c(\mathbf{z}_p^{\mathbb{H}}, \mathbf{z}_{f^+}^{\mathbb{H}}) 
 - d_{\mathbb{H}}^c(\mathbf{z}_p^{\mathbb{H}}, \mathbf{z}_{f^-}^{\mathbb{H}}) + \gamma \Big]_+,
\end{align}
where $d_{\mathbb{H}}^c(\cdot,\cdot)$ denotes hyperbolic distance and $[\cdot]_+ = \max(0, \cdot)$ ensures non-negative loss.

\paragraph{Fact-to-Passage alignment}
Symmetrically, for each fact $f \in \mathcal{F}$, let $\mathcal{P}(f)$ denote its supporting passages. Each positive pair $(f, p^+)$ is contrasted with a negative passage $p^-$ not containing $f$, enforcing the reverse containment:
\begin{align}
\mathcal{L}_{f \to p}
& = \sum_{p \in \mathcal{P}(f)}
\Big[ d_{\mathbb{H}}^c(\mathbf{z}_f^{\mathbb{H}}, \mathbf{z}_{p^+}^{\mathbb{H}}) - d_{\mathbb{H}}^c(\mathbf{z}_{f}^{\mathbb{H}}, \mathbf{z}_{p^-}^{\mathbb{H}}) + \gamma \Big]_+.
\end{align}

The dual alignment jointly optimizes hierarchical consistency in both directions: passages act as semantic containers that aggregate multiple fine-grained facts, while facts serve as evidence grounding for passages. This bidirectional constraint stabilizes the learned geometry and prevents degenerate alignment (e.g., all nodes collapsing toward the boundary). 

\subsection{Dual-space Retrieval Process}
Building upon the retrieval framework of~\cite{gutierrez2025rag}, we extend it to a dual-space setting that jointly exploits the complementary strengths of Euclidean and hyperbolic geometries. Specifically, our retrieval module introduces two key enhancements:
1) it performs independent relevance propagation in both the \emph{Euclidean space}, which excels at modeling local semantic similarity and the \emph{Hyperbolic space} which naturally preserves hierarchical containment;
2) it integrates their results through a mutual-ranking fusion strategy that emphasizes cross-space consistency while preventing interference during propagation.

\begin{figure*}[htbp]
    \centering
    \includegraphics[width=\linewidth]{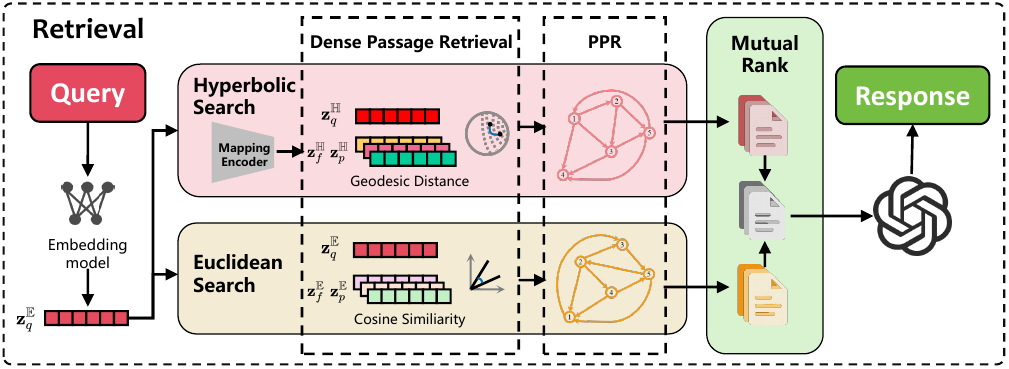}
    \caption{\textbf{Illustration of the dual-space retrieval framework.} The query is processed in parallel Euclidean and hyperbolic spaces.
Each branch computes query–fact similarities through different ways, propagates them to entities, and combines them with direct query–passage priors to form a seed distribution for PPR on the passage–entity graph, yielding space-specific rankings ($\mathcal{R}^{\mathbb{E}}$ and $\mathcal{R}^{\mathbb{H}}$). A mutual-ranking fusion then favors passages consistently ranked high in both spaces, balancing Euclidean semantic similarity and hyperbolic hierarchical structure for robust retrieval.}
    \label{fig:retrieval}
\end{figure*}

This dual-space design enables retrieval to simultaneously benefit from fine-grained topical alignment and geometry-aware reasoning over hierarchical relations. An overview of the complete dual-space retrieval workflow is shown in Fig.~\ref{fig:retrieval}

\subsubsection{Euclidean branch (semantic similarity-based retrieval)}  
The Euclidean branch aims to model fine-grained semantic similarity between the query and corpus elements. Its operation proceeds in three stages.

\paragraph{Signal initialization}
Given a query $q$, we encode it into the Euclidean space using the same pretrained encoder $\mathrm{Enc}(\cdot)$ as in the indexing stage, obtaining the query embedding $\mathbf{z}_q^{\mathbb{E}}$.
Two complementary types of initial relevance signals are derived:

\begin{itemize}
\item \textbf{Fact-level signals.}
We compute cosine similarity between $\mathbf{z}_q^{\mathbb{E}}$ and cached fact embeddings $\mathbf{z}_f^{\mathbb{E}}$ from the index. The top-$k$ most similar facts are selected as initial evidence. Their scores are propagated to the corresponding subject and object entities ($e_s, e_o$) and normalized by each entity’s number of associated passages to mitigate degree bias.
\item \textbf{Passage-level signals.}
We directly compute cosine similarity between $\mathbf{z}_q^{\mathbb{E}}$ and passage embeddings $\mathbf{z}_p^{\mathbb{E}}$, forming passage-level priors that capture topical alignment.
\end{itemize}

\paragraph{Seed distribution construction}
The signals are merged into a unified seed distribution $\mathbf{s}_q^{\mathbb{E}}$, where each entry corresponds to the initial relevance weight of a node in the heterogeneous graph. This distribution serves as the restart distribution for the Personalized Page Rank (PPR) ~\cite{bahmani2010fast} process, ensuring propagation is centered on query-relevant regions.

\paragraph{Graph propagation}
% We apply PPR over the passage–entity graph using $\mathbf{s}_q^{\mathbb{E}}$ as the seed vector:
% \begin{equation}
% \pi_q^{\mathbb{E}} = \alpha\mathbf{s}_q^{\mathbb{E}} + (1-\alpha)\pi_q^{\mathbb{E}} W,
% \end{equation}
% where $W$ is the row-normalized adjacency matrix and $\alpha \in (0,1)$ is the restart probability controlling the balance between local focus and global diffusion.
We apply PPR over the passage–entity graph using $\mathbf{s}_q^{\mathbb{E}}$ as the seed vector, computing 
$\pi_q^{\mathbb{E}} = \alpha \mathbf{s}_q^{\mathbb{E}} + (1 - \alpha)\,\pi_q^{\mathbb{E}} W$, 
where $W$ is the row-normalized adjacency matrix and $\alpha \in (0,1)$ is the restart probability controlling the balance between local focus and global diffusion.
After convergence, the stationary distribution $\pi_q^{\mathbb{E}}$ yields passage-level relevance scores, which are sorted to form the Euclidean ranking list $\mathcal{R}^{\mathbb{E}}$.

\subsubsection{Hyperbolic branch (hierarchy-aware retrieval)}
The hyperbolic branch follows the same computational structure as the Euclidean branch, but performs operations in a non-Euclidean manifold, thereby contributing complementary hierarchical signals.

\paragraph{Query projection and signal initialization}
The Euclidean query embedding $\mathbf{z}_q^{\mathbb{E}}$ is projected into the hyperbolic space $\mathbb{H}_d^c$ via the trained projection module (Section~\ref{etoh}), producing $\mathbf{z}_q^{\mathbb{H}}$.

\begin{itemize}
\item \textbf{Fact-level signals.}
We measure similarity as the \textit{negative hyperbolic geodesic distance} between $\mathbf{z}_q^{\mathbb{H}}$ and cached fact embeddings $\mathbf{z}_f^{\mathbb{H}}$, selecting the top-$k$ closest facts as initial evidence. Their scores are propagated to associated entities ($e_s$, $e_o$) and normalized by entity degree, identical to the Euclidean branch.
\item \textbf{Passage-level signals.}
Similarly, negative hyperbolic distance between $\mathbf{z}_q^{\mathbb{H}}$ and passage embeddings $\mathbf{z}_p^{\mathbb{H}}$ provides hierarchy-aware passage priors, emphasizing fine-to-coarse structural proximity.
\end{itemize}

\paragraph{Seed distribution and propagation.}
% The entity- and passage-level signals are merged into a hyperbolic seed distribution $\mathbf{s}_q^{\mathbb{H}}$, which serves as the restart distribution for the PPR process:
% \begin{equation}
% \pi_q^{\mathbb{H}} = \alpha\mathbf{s}_q^{\mathbb{H}} + (1-\alpha)\pi_q^{\mathbb{H}} W,
% \end{equation}
% where $W$ and $\alpha$ are shared with the Euclidean branch to maintain propagation consistency.
The entity and passage-level signals are merged into a hyperbolic seed distribution $\mathbf{s}_q^{\mathbb{H}}$, which serves as the restart distribution for the PPR process, yielding 
$\pi_q^{\mathbb{H}} = \alpha \mathbf{s}_q^{\mathbb{H}} + (1 - \alpha)\,\pi_q^{\mathbb{H}} W$, 
where $W$ and $\alpha$ are shared with the Euclidean branch to ensure consistent propagation dynamics.
After convergence, the stationary distribution $\pi_q^{\mathbb{H}}$ defines the hyperbolic relevance scores for passages, producing the ranking list $\mathcal{R}^{\mathbb{H}}$.

\subsubsection{Mutual-ranking fusion (integrating complementary signals)}  
To combine the Euclidean ($\mathcal{R}^{\mathbb{E}}$) and hyperbolic ($\mathcal{R}^{\mathbb{H}}$) rankings, we employ a mutual-ranking fusion scheme that emphasizes passages consistently favored by both spaces, which mitigates noise from either single space and amplifies robust signals. The fusion process has three key steps:
\begin{enumerate}
    \item \textbf{Reciprocal rank calculation.} For each passage $p$, we convert its rank in each list into a reciprocal-rank score, 
$s_{\mathbb{E}}(p)=1/(\mathrm{rank}_{\mathbb{E}}(p)+1)$ and 
$s_{\mathbb{H}}(p)=1/(\mathrm{rank}_{\mathbb{H}}(p)+1)$, 
where $\mathrm{rank}_{\mathbb{E}}(p)$ and $\mathrm{rank}_{\mathbb{H}}(p)$ denote the rank of $p$ in $\mathcal{R}^{\mathbb{E}}$ and $\mathcal{R}^{\mathbb{H}}$, respectively.

   % The $+1$ term avoids division by zero for the top-ranked passage.
   \item \textbf{Consistency bonus calculation.} We assign an additional bonus $b(p)$ to passages that appear in both rankings, rewarding cross-space consistency. The bonus is computed as $b(p)=1/(\mathrm{rank}_{\mathbb{E}}(p)+\mathrm{rank}_{\mathbb{H}}(p)+2)$, which gives higher values to passages that simultaneously achieve strong ranks in both lists.

   % where the $+2$ term ensures the denominator is positive even if a passage is the top-ranked in both lists.
   \item \textbf{Hybrid score computation.} The final hybrid score for each passage $p$ is computed as 
$s_{\text{hyb}}(p)=\big(s_{\mathbb{E}}(p)+s_{\mathbb{H}}(p)\big)\,(1+b(p))$, 
and passages are subsequently re-ranked in descending order of $s_{\text{hyb}}(p)$ to obtain the final retrieval result.

\end{enumerate}
This late-fusion design ensures that Euclidean and hyperbolic retrieval remain independent during graph propagation, while the mutual-ranking scheme explicitly leverages cross-space consistency to enhance retrieval robustness and precision.

\section{Evaluation}\label{experiments}
\begin{table*}[tbp]
\caption{Summary of dataset information, extraction results, and graph statistics.}
\label{tab:dataset_stats}
\centering
% \footnotesize
% \resizebox{\linewidth}{!}{
% \scalebox{1.5}{
\setlength{\tabcolsep}{25pt}
\begin{tabular}{@{}cccccc@{}}
\toprule
                    & NQ         & PopQA      & MuSiQue    & 2Wiki    & HotpotQA    \\ \midrule
\multicolumn{6}{c}{\cellcolor[HTML]{E6E6E6}\textit{\textbf{Basic Dataset Statistic}}}                \\
Number of Queries      & 1,000       & 1,000       & 1,000       & 1,000     & 1,000        \\
Number of Passages     & 9,633       & 8,676       & 11,656      & 6,119     & 9,811        \\ \midrule
\multicolumn{6}{c}{\cellcolor[HTML]{E6E6E6}\textit{\textbf{Information Extraction Statistic}}} \\
Number of Facts        & 115,243     & 112,990     & 140,739     & 68,840    & 129,997      \\
Number of Entities     & 62,234      & 72,050      & 85,274      & 44,003    & 81,200       \\ \midrule
\multicolumn{6}{c}{\cellcolor[HTML]{E6E6E6}\textit{\textbf{Knowledge Graph Statistic}}} \\
Number of Nodes        & 71,867     & 80,726     & 96,944     & 50,123    & 91,011      \\
Number of Edges     & 990,057      & 954,528      & 1,399,262      & 726,330    & 1,246,677       \\ 
\bottomrule
\end{tabular}
% }
\end{table*}

\subsection{Datasets}

To evaluate the effectiveness of our dual-space retrieval framework in supporting multi-hop reasoning, we follow existing work \cite{gutierrez2025rag} categorizing datasets into two challenge types:
\begin{enumerate}
    \item \textbf{Simple QA.} This category primarily evaluates the ability to recall and retrieve factual knowledge accurately. We randomly select 1,000 queries from Natural Questions (NQ) \cite{wang2024rear}, which contains real user questions covering diverse topics. Additionally, we select 1,000 queries from PopQA \cite{mallen2022not}, derived from the December 2021 Wikipedia dump. Both datasets provide straightforward QA pairs suitable for assessing single-hop retrieval performance. Notably, PopQA is more entity-centric, with entities occurring less frequently than in NQ, making it particularly useful for evaluating entity recognition and retrieval in simple QA tasks.

    \item \textbf{Multi-hop QA.} Multi-hop datasets require the model to connect multiple pieces of information to answer a query, testing associative reasoning capabilities. We sample 1,000 queries from MuSiQue \cite{trivedi2022musique}, 2WikiMultihopQA \cite{ho2020constructing}, and HotpotQA \cite{yang2018hotpotqa}, following the setup in HippoRAG \cite{gutierrez2025rag}. For all multi-hop datasets, long-form contexts are segmented into shorter passages while maintaining the same RAG setup, allowing our retrieval framework to aggregate evidence across multiple passages.
\end{enumerate}

The statistics of the sampled datasets are summarized in Table \ref{tab:dataset_stats}. Together, these datasets provide a comprehensive evaluation of retrieval models on factual memory, multi-hop reasoning, and discourse-level comprehension.

\subsection{Baselines}
We evaluate our framework against three categories of baselines:
\begin{enumerate}
    \item \textbf{Simple retrieval methods.} BM25 \cite{robertson1994some}: A classical lexical matching baseline. Contriever \cite{izacard2021unsupervised} and GTR \cite{ni2021large}: Popular dense embedding retrievers that rely solely on Euclidean semantic similarity.
    \item \textbf{Large pre-trained embedding models.} These baselines use state-of-the-art 7B-scale embedding models that achieve strong performance on the BEIR benchmark \cite{thakur2021beir}: Alibaba-NLP/GTE-Qwen2-7B-Instruct \cite{li2023towards}, GritLM/GritLM-7B \cite{Muennighoff2024GenerativeRI}, nvidia/NV-Embed-v2 \cite{lee2024nv}. They provide strong semantic representations and serve as a competitive reference for dense retrieval performance.
    \item \textbf{Structure-augmented RAG.} These methods leverage graph or hierarchical structures to improve multi-hop reasoning: RAPTOR \cite{sarthi2024raptor}: Organizes the corpus hierarchically based on semantic similarity. GraphRAG \cite{edge2024local} and LightRAG \cite{guo2024lightrag}: Use knowledge graphs to propagate relevance and summarize high-level concepts. HippoRAG \cite{jimenez2024hipporag}: Integrates graph-based knowledge using PPR rather than summarization. HippoRAG2 \cite{gutierrez2025rag}: An improved variant of HippoRAG that refines both graph-based retrieval and memory aggregation, yielding stronger performance on multi-hop QA benchmarks.
\end{enumerate}

\subsection{Metrics}
We adopt two complementary sets of metrics to evaluate retrieval and downstream QA performance.
\begin{itemize}
    \item \textbf{Retrieval evaluation.} Following HippoRAG (Gutiérrez et al., 2024), we report Passage Recall@5, which measures whether the gold evidence passages appear among the top-$5$ retrieved candidates. 
    \item \textbf{QA evaluation.} For end-to-end question answering, we adopt the token-level evaluation protocol introduced in MuSiQue \cite{trivedi2022musique}. We report both the Exact Match (EM) and F1 scores between the predicted answer span and the ground-truth answer. EM measures the exact string match accuracy, while F1 balances precision and recall by capturing the overlap of tokens between prediction and reference.
\end{itemize}
Together, these metrics provide a comprehensive view: retrieval recall emphasizes evidence coverage, while EM and F1 measures final answer quality.

\subsection{Implementation Details}
We follow the experimental setup of HippoRAG2 \cite{gutierrez2025rag} to ensure fair comparison. Specifically, we use Llama-3.3-70B-Instruct~\cite{llama3} as the extraction model for both NER and OpenIE, and as the triple filtering model. For retrieval, we adopt NV-Embed-v2 \cite{lee2024nv} as the embedding model. For QA generation, we report the results of feeding the top-5 retrieved passages as context to an LLM (i.e., Llama-3.3-70B-Instruct). All hyperparameters (e.g., damping factor in PPR, retrieval cutoffs) follow the default values from HippoRAG2 unless otherwise stated.

\subsection{Experimental Results}
We evaluate HyperbolicRAG through a comprehensive set of experiments to assess its retrieval effectiveness, end-to-end QA performance, component-wise contributions, and model-agnostic robustness. 
Across all results, our method is highlighted in \colorbox{lightgraycell}{gray}, the best result is marked in \textbf{bold}, and the second-best result is \underline{underlined}.

\subsubsection{Information Extraction Results and Graph Statistics}
Before evaluating retrieval performance, we first report the information extraction and graph construction results from the indexing stage. Each passage is converted into structured fact triples using Llama-3.3-70B-Instruct, forming the factual backbone of the heterogeneous passage–entity graph. Based on these facts and entities, we construct the passage–entity knowledge graphs, where passages and entities serve as nodes and edges represent factual or co-occurrence relations. The overall extraction results and graph statistics are summarized in Table~\ref{tab:dataset_stats}.
\begin{table*}[t]
\centering
\caption{Comparison of retrieval methods in terms of Recall@5 (\%) across simple QA and multi-hop QA datasets.}
\label{tab:recall@5}
% \footnotesize
% \resizebox{0.9\linewidth}{!}{
\setlength{\tabcolsep}{18pt}
\begin{tabular}{ccccccc}
\toprule
                                             & \multicolumn{2}{c}{\textbf{Simple QA}} & \multicolumn{3}{c}{\textbf{Multi-Hop QA}} &                                \\ 
\cmidrule(lr){2-6}
\multirow{-2}{*}{\textbf{Retrieval Methods}} & NQ                 & PopQA             & MuSiQue      & 2Wiki      & HotpotQA      & \multirow{-2}{*}{\textbf{Avg}} \\ 
\midrule

% ---------- Simple Baselines ----------
\multicolumn{7}{c}{\textbf{\textit{Simple Baselines}}} \\[-2pt]
\midrule
BM25                                         & 56.1               & 35.7              & 43.5         & 65.3       & 74.8          & 55.1                           \\
Contriever                                   & 54.6               & 43.2              & 46.6         & 57.5       & 75.3          & 55.4                           \\
GTR (T5-base)                                & 63.4               & 49.4              & 49.1         & 67.9       & 73.9          & 60.7                           \\[3pt] 
\midrule

% ---------- Large Embedding Models ----------
\multicolumn{7}{c}{\textbf{\textit{Large Embedding Models}}} \\[-2pt]
\midrule
GTE-Qwen2-7B-Instruct                        & 74.3               & 50.6              & 63.6         & 74.8       & 89.1          & 70.5                           \\
GritLM-7B                                    & 76.6               & 50.1              & 65.9         & 76.0       & 92.4          & 72.2                           \\
NV-Embed-v2-7B                               & 75.4               & 51.0              & 69.7         & 76.5       & 94.5          & 73.4                           \\[3pt] 
\midrule

% ---------- Structure-Augmented RAG ----------
\multicolumn{7}{c}{\textbf{\textit{Structure-Augmented RAG}}} \\[-2pt]
\midrule
RAPTOR                                       & 68.3               & 48.7              & 57.8         & 66.2       & 86.9          & 65.6                           \\
HippoRAG                                     & 44.4               & \textbf{53.8}     & 53.2         & 90.4       & 77.3          & 63.8                           \\
HippoRAG2                                    & \underline{78.0}   & 51.7              & \underline{74.7} & \underline{90.4} & \underline{96.2} & 78.2                           \\[1pt]
\midrule
\rowcolor[HTML]{E6E6E6}
\textbf{HyperbolicRAG}                       & \textbf{78.5}      & \underline{51.9}  & \textbf{76.2} & \textbf{92.1} & \textbf{96.3} & \textbf{79.0}                  \\ 
\bottomrule
\end{tabular}

% }
\end{table*}

\subsubsection{Retrieval Results}

We evaluate retrieval performance using Recall@5. Table~\ref{tab:recall@5} reports the results on both simple and multi-hop QA datasets. HyperbolicRAG achieves the highest overall Recall@5 of 79.0\%, outperforming all Euclidean and structure-augmented baselines. Compared with the strongest Euclidean retriever, NV-Embed-v2-7B (73.4\%), it delivers a 5.6\% absolute improvement, demonstrating the advantage of modeling hierarchical organization beyond surface-level similarity. Within structure-augmented methods, HyperbolicRAG slightly outperforms HippoRAG2 (78.2\%), indicating that hyperbolic geometry provides complementary benefits even for advanced graph-based retrieval frameworks. The gains are most evident on multi-hop datasets such as 2Wiki (92.1\% vs.\ 90.4\%) and MuSiQue (76.2\% vs.\ 74.7\%), where reasoning requires integrating multiple entities and relations. On simpler datasets such as NQ and PopQA, improvements are smaller, confirming that the hyperbolic formulation enhances robustness without overfitting to specific structural patterns. Overall, these findings demonstrate that explicitly modeling relational hierarchies in hyperbolic space mitigates the hubness bias inherent in Euclidean embeddings and leads to more precise and context-aware retrieval.

\begin{table*}[t]
\centering
\caption{EM and F1 (\%) performance comparison of retrieval methods using the top-5 retrieved passages.}
\label{tab:F1_5}
% \footnotesize
% \resizebox{\linewidth}{!}{
\setlength{\tabcolsep}{18pt}
\begin{tabular}{ccccccc}
\toprule
                                             & \multicolumn{2}{c}{\textbf{Simple QA}} & \multicolumn{3}{c}{\textbf{Multi-Hop QA}} &                                \\ 
\cmidrule(lr){2-6}
\multirow{-2}{*}{\textbf{Retrieval Methods}} & NQ                 & PopQA             & MuSiQue      & 2Wiki      & HotpotQA      & \multirow{-2}{*}{\textbf{Avg}} \\ 
\midrule

% ---------- Simple Baselines ----------
\multicolumn{7}{c}{\textbf{\textit{Simple Baselines}}} \\[-2pt]
\midrule
None                                         & 40.2/54.9               & 28.2/32.5              & 17.6/26.1         & 36.5/42.8       & 37.0/47.3          & 31.9/40.7                           \\
BM25                                         & 45.0/58.9              & 41.6/53.1              & 24.0/31.3         & 38.1/41.9       & 51.3/62.3          &  40.0/49.5                          \\
Contriever                                   & 44.7/59.0             & 39.1/49.9              & 20.3/28.8         & 47.9/51.2       & 52.0/63.4          &  40.8/50.5                          \\
GTR (T5-base)                                & 45.5/59.9               & \underline{43.2}/56.2              & 25.8/34.6         & 49.2/52.8       & 50.6/62.8          &  42.8/53.3                          \\[3pt]
\midrule

% ---------- Large Embedding Models ----------
\multicolumn{7}{c}{\textbf{\textit{Large Embedding Models}}} \\[-2pt]
\midrule
GTE-Qwen2-7B-Instruct                        & 46.6/62.0               & \textbf{43.5}/\underline{56.3}             & 30.6/40.9         & 55.1/60.0       & 58.6/71.0 & 46.9/58.0                           \\
GritLM-7B                                    & 46.8/61.3               & 42.8/55.8              & 33.6/44.8         & 55.8/60.6       & 60.7/73.3          &  47.9/59.2                        \\
NV-Embed-v2-7B                               & \underline{47.3}/61.9               & 42.9/55.7              & 34.7/45.7         & 57.5/61.5       & \textbf{62.8}/\underline{75.3}          &  49.0/60.0                         \\[3pt]
\midrule

% ---------- Structure-Augmented RAG ----------
\multicolumn{7}{c}{\textbf{\textit{Structure-Augmented RAG}}} \\[-2pt]
\midrule
RAPTOR                                       & 36.9/50.7               & 43.1/56.2              & 20.7/28.9         & 47.3/52.1       & 56.8/69.5          &   40.9/51.5                       \\
GraphRAG                                     & 30.8/46.9               & 31.4/48.1              & 27.3/38.5         & 51.4/58.6       & 55.2/68.6          &   39.2/52.1                       \\
LightRAG                                     & 8.6/16.6                & 2.1/2.4                & 0.5/1.6           & 9.4/11.6        & 2.0/2.4            &  4.5/6.9                         \\
HippoRAG                                     & 43.0/55.3               & 42.7/55.9              & 26.2/35.1         & 65.0/71.8       & 52.6/63.5          & 45.9/56.3                          \\
HippoRAG2                                    & 47.1/\textbf{62.0}      & 42.9/56.2              & 37.2/48.6         & \underline{65.0}/\underline{71.0} & \underline{62.7}/\textbf{75.5} & \underline{51.0}/\underline{62.7} \\[-1pt]
\midrule
\rowcolor[HTML]{E6E6E6}
\textbf{HyperbolicRAG}                       & \textbf{47.8/62.3}      & 42.4/\textbf{56.3}     & \textbf{39.5/50.6} & \textbf{65.5/72.3} & 61.7/75.2         &  \textbf{51.4/63.3}              \\ 
\bottomrule
\end{tabular}
% }
\end{table*}

\begin{table}[t]
\centering
\footnotesize
\caption{Ablation study on multi-hop QA datasets (Recall@5). }
\label{tab:ablation}
\resizebox{\linewidth}{!}{
\begin{tabular}{c|ccc}
\toprule
                                            & \multicolumn{3}{c}{\textbf{Multi-Hop QA Dataset}}                                               \\ \cmidrule(l){2-4} 
\multirow{-2}{*}{\textbf{Retrieval Methods}} & \multicolumn{1}{c|}{\textbf{Musique}} & \multicolumn{1}{c|}{\textbf{2Wiki}} & \textbf{HotpotQA} \\ \midrule
Euclidean Alignment                         & 71.4                                  & 88.4                                & 94.8              \\ \midrule
HyperbolicRAG  w/o Hyperbolic Signal          & 74.7                                  & 90.4                                & 96.2              \\ \midrule
HyperbolicRAG  w/o Euclidean Signal           & 73.9                                  & 90.4                                & 95.9              \\ \midrule
\rowcolor[HTML]{E6E6E6} 
\textbf{HyperbolicRAG}                      & \textbf{76.2}                         & \textbf{91.1}                       & \textbf{96.3}     \\ \bottomrule
\end{tabular}
}
\end{table}
\subsubsection{Generation Results}
Beyond retrieval effectiveness, we further evaluate end-to-end QA performance using EM and token-level F1 on answers generated from the top five retrieved passages. Table~\ref{tab:F1_5} summarizes the results. HyperbolicRAG achieves the highest overall performance, with an average of 51.4\% EM and 63.3\% F1, outperforming both Euclidean and structure-augmented baselines. Compared with the strongest competitors, HippoRAG2 (51.0\% / 62.7\%) and NV-Embed-v2-7B (49.0\% / 60.0\%), HyperbolicRAG demonstrates consistent gains across all datasets. The improvement is particularly pronounced on multi-hop QA benchmarks such as MuSiQue (39.5\% / 50.6\%) and 2Wiki (65.5\% / 72.3\%), where reasoning requires integrating multiple entities and factual relations. In these tasks, hyperbolic representations capture more coherent and hierarchically consistent evidence, enabling the generator to produce more complete and faithful answers. On simpler datasets such as NQ and PopQA, the model maintains competitive results (47.8\% / 62.3\% and 42.4\% / 56.3\%), indicating that curvature-based modeling preserves generalization on non-compositional queries.
Overall, these findings show that HyperbolicRAG provides the LLM with more precise and semantically grounded evidence, leading to higher factual consistency and completeness in generated responses.

\subsubsection{Ablation Results}
To examine the contribution of each component in HyperbolicRAG, we conduct ablation studies on three representative multi-hop QA datasets, as shown in Table~\ref{tab:ablation}.
The first variant, Euclidean Embedding Alignment, replaces the hyperbolic manifold with a flat Euclidean space while retaining the same contrastive learning objective. Although the pull–push optimization encourages hierarchical alignment, its performance (71.4\% on MuSiQue, 88.4\% on 2Wiki, and 94.8\% on HotpotQA) falls short. This degradation reflects an inherent limitation of Euclidean space, whose isotropic geometry captures pairwise similarity but fails to express asymmetric containment among passages, entities, and facts. Consequently, hierarchical signals are compressed into a single representational layer, reducing the model’s ability to distinguish between abstract and specific evidence.

The second and third variants assess the impact of the dual-space fusion mechanism by disabling one of the ranking channels. Without hyperbolic ranking signal, the model relies purely on Euclidean similarity (74.7\% on MuSiQue and 90.4\% on 2Wiki); conversely, removing the Euclidean ranking yields a comparable decline (73.9\% and 90.4\%). These results suggest that the two spaces capture complementary aspects of relevance: the Euclidean space refines local semantic consistency, whereas the hyperbolic space preserves global structural hierarchy. The complete HyperbolicRAG achieves the best overall performance (76.2\%, 91.1\%, and 96.3\%), confirming that the proposed rank-level fusion combines hyperbolic and Euclidean signals to deliver semantically precise and structurally coherent retrieval.
\begin{figure}[t]
    \centering
    \includegraphics[width=\linewidth]{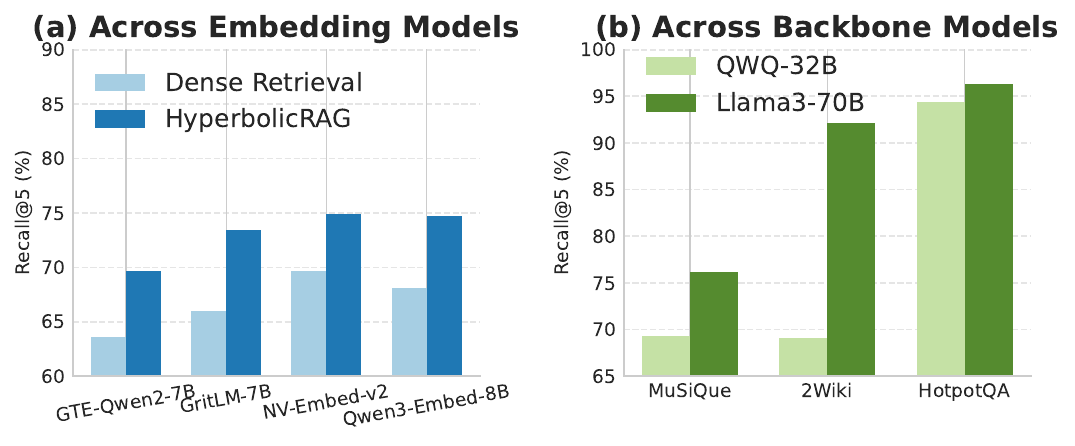}
    \caption{Comparison of HyperbolicRAG under (a) different embedding encoders and (b) generative backbones.}
    \label{fig:dense_retriever}
\end{figure}
\subsubsection{Model-Agnostic Effectiveness}

We assess the compatibility of HyperbolicRAG with a broad set of dense encoders and LLM backbones. As shown in Fig.~\ref{fig:dense_retriever}(a), integrating our hierarchical enhancement mechanism with various dense retrievers, including GTE Qwen2 7B Instruct, GritLM 7B, NV Embed v2 (7B), and Qwen3 Embedding 8B~\cite{zhang2025qwen3}, consistently yields higher Passage Recall@5 compared with their Euclidean counterparts. These results indicate that the hyperbolic representation reliably preserves hierarchical semantics across diverse embedding distributions.

In addition to retrieval encoders, Fig.~\ref{fig:dense_retriever}(b) demonstrates that HyperbolicRAG provides stable performance gains when paired with different LLM backbones. Although Llama3 70B serves as our primary generator, comparable improvements are observed with QWQ 32B~\cite{qwq32b}, which confirms that the advantages of hierarchical enhancement generalize across architectures of different capacities. We also find that the overall retrieval quality is influenced by the accuracy of the relational graph construction pipeline. Inaccurate or incomplete extraction of entities and relations may introduce noise and fragmentation, which in turn limits the achievable performance. Despite these constraints, HyperbolicRAG consistently enhances retrieval robustness and hierarchical sensitivity across both retriever and backbone variations.

\subsubsection{Effect of the Curvature Hyperparameter}
\begin{figure}[t]
    \centering
    \includegraphics[width=\linewidth]{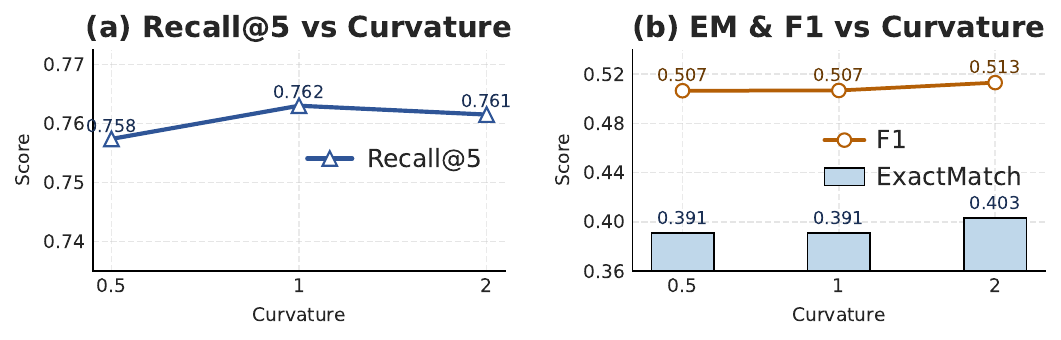}
    \caption{Effect of the curvature hyperparameter on retrieval and generation performance.}
    \label{fig:curvature}
\end{figure}

To assess the sensitivity of the model to geometric settings, we vary the curvature hyperparameter $c$.
As illustrated in Fig.~\ref{fig:curvature}, 
$c$ has minimal influence on retrieval performance, while a moderate curvature leads to slight but consistent improvements in generation metrics.
This suggests that the model is robust to curvature variations and benefits marginally from non-Euclidean geometry.

\section{Conclusion}\label{conclusion}
In this work, we propose HyperbolicRAG, a hierarchy-aware retrieval framework that captures the intrinsic structure of entity–fact–passage relations through hyperbolic geometry.
By modeling hierarchical containment within a curved space, HyperbolicRAG effectively alleviates the hubness bias inherent in Euclidean embeddings and improves the precision of evidence retrieval.
A dual-space retrieval mechanism further integrates Euclidean and hyperbolic reasoning, combining fine-grained semantic similarity with global structural awareness.
Extensive experiments across multiple QA benchmarks demonstrate consistent gains in both retrieval and answer generation, particularly on multi-hop reasoning tasks.

\bibliographystyle{IEEEtran}
\bibliography{ref}

% \vfill

\end{document}